\documentclass[10pt,twocolumn,letterpaper]{article}

\usepackage{cvpr}              
\definecolor{cvprblue}{rgb}{0.21,0.49,0.74}
\usepackage[pagebackref,breaklinks,colorlinks,allcolors=cvprblue]{hyperref}

\usepackage{graphicx}
\usepackage{multirow}
\usepackage{pifont}
\usepackage{fontawesome5}
\usepackage{booktabs}
\usepackage{xcolor}
\usepackage[accsupp]{axessibility}

\definecolor{alizarin}{rgb}{0.82, 0.1, 0.26}
\definecolor{celadon}{rgb}{0.67, 0.88, 0.69}
\definecolor{junglegreen}{rgb}{0.16, 0.67, 0.53}

\usepackage[capitalize]{cleveref}
\crefname{section}{Sec.}{Secs.}
\Crefname{section}{Section}{Sections}
\Crefname{table}{Table}{Tables}
\crefname{table}{Tab.}{Tabs.}

\newcommand{\method}{MoonSeg3R\xspace}
\def\paperID{*****} 
\def\confName{CVPR}
\def\confYear{2026}

\title{\method: Monocular Online Zero-Shot Segment Anything in 3D with Reconstructive Foundation Priors}

\author{Zhipeng Du\textsuperscript{1} \hspace{0.8em}  Duolikun Danier\textsuperscript{1} \hspace{0.8em}  Jan Eric Lenssen\textsuperscript{2}  \hspace{0.8em}  Hakan Bilen\textsuperscript{1}\\
{\tt\small \{zhipeng.du,duolikun.danier,h.bilen\}@ed.ac.uk \quad jlenssen@mpi-inf.mpg.de}\vspace{0.2cm}\\
 \textsuperscript{1}University of Edinburgh \quad \textsuperscript{2}Max Planck Institute for Informatics, SIC\\
}

\begin{document}
\maketitle
\begin{abstract}

In this paper, we focus on online zero-shot monocular 3D instance segmentation, a novel practical setting where existing approaches fail to perform because they rely on posed RGB-D sequences. To overcome this limitation, we leverage CUT3R, a recent Reconstructive Foundation Model (RFM), to provide reliable geometric priors from a single RGB stream. 
We propose \method, which introduces three key components: (1) a self-supervised query refinement module with spatial–semantic distillation that transforms segmentation masks from 2D visual foundation models (VFMs) into discriminative 3D queries; (2) a 3D query index memory that provides temporal consistency by retrieving contextual queries; and (3) a state-distribution token from CUT3R that acts as a mask identity descriptor to strengthen cross-frame fusion. Experiments on ScanNet200 and SceneNN show that MoonSeg3R is the first method to enable online monocular 3D segmentation and achieves performance competitive with state-of-the-art RGB-D–based systems. Our code is available at \small{\url{https://github.com/VICO-UoE/MoonSeg3R}}.

\end{abstract}    
\section{Introduction}
\label{sec:intro}

\begin{figure}[!t]
    \centering
    \includegraphics[width=\linewidth]{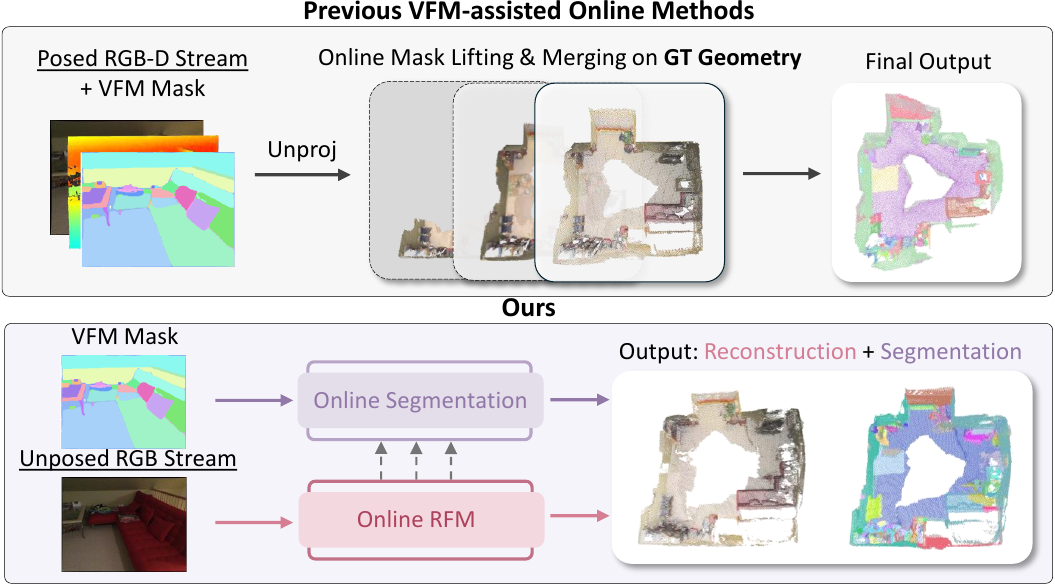}
    \caption{Previous VFM-assisted Online Paradigm \emph{v.s.} Ours. While existing methods rely on the ground truth geometry (and 3D segmentation masks), our method works in a monocular online zero-shot setting, exploiting the spatio-temporal priors from an RFM to help with online 3D segmentation, thereby simultaneously achieving online reconstruction and segmentation.}
    \label{fig:teaser}
    \vspace{-10px}
\end{figure}

Monocular online 3D instance segmentation, the task of incrementally reconstructing and segmenting 3D object instances from a streaming RGB sequence, represents a key capability for embodied perception and autonomous operation in complex real-world environments such as robotic navigation and interaction~\cite{chaplot2020nips,mousavian2019iccv,zhang2023cvpr}. 
Unlike offline methods, the online setting demands maintaining temporally consistent geometry and semantics from a single camera stream, in real time, and without explicit 3D supervision. 
This makes the task especially challenging: observations are partial, geometry must be inferred implicitly, and 3D instance associations must be maintained despite view changes and occlusions.

Recent works leverage Visual Foundation Models (VFMs) such as SAM~\cite{kirillov2023sam} and CLIP~\cite{radford2021icml} to provide powerful 2D mask priors, which can then be lifted into 3D using explicit geometry from depth sensors. 
Methods such as EmbodiedSAM~\cite{embodiedsam2025iclr} and OnlineAnySeg~\cite{tang2025onlineanyseg} have demonstrated promising online 3D segmentation performance by combining VFM-based semantic cues with depth information for mask lifting and merging, either in a fully-supervised or zero-shot manner. 
However, these methods still require accurate depth or point cloud supervision, which limits deployment on platforms where dedicated depth sensors are unavailable or impractical~\cite{wu2024panorecon}. 
A fundamental open question remains: \emph{Can we perform online 3D instance segmentation directly from monocular RGB inputs, without ground-truth geometry or instance masks?}

Recent feed-forward online reconstruction networks~\cite{wang2024dust3r,wang2025cut3r,zhang2025monst3r,wang2025vggt}, termed \textit{Reconstructive Foundation Models (RFMs)}, enable real-time, generalizable 3D reconstruction from unposed monocular streams. 
Among them, CUT3R~\cite{wang2025cut3r} enables online inference capabilities by maintaining a recurrent \emph{state token} that implicitly encodes global scene geometry. 
Nevertheless, its representations are optimized for reconstruction rather than segmentation, yielding three key limitations: (i) lack of object-level semantic awareness, (ii) noisy explicit geometry predictions, and (iii) a latent memory that is powerful yet non-interpretable for downstream perception tasks.

We introduce \textbf{\method}, a monocular, online, zero-shot 3D instance segmentation framework that integrates the geometric priors of reconstructive foundation models (RFMs) with the semantic power of visual foundation models (VFMs), as shown in Fig.~\ref{fig:teaser}. 
\method reconstructs the scene and segments instances online by unprojecting VFM masks into 3D and associating them over time. 
It comprises four key components: (i) a \textbf{query refinement module} that transforms 2D masks into discriminative 3D prototype queries via cross-attention with geometric and semantic features, (ii) a \textbf{spatial-semantic distillation} objective for self-supervised query training, (iii) a \textbf{3D Query Index Memory (QIM)} for temporally consistent query retrieval and cross-frame association, and (iv) a \textbf{state distribution token} that captures CUT3R’s attention dynamics to support robust mask fusion. 
These modules enable consistent 3D segmentation purely from monocular RGB streams, without depth or mask supervision.
We evaluate \method on \textbf{ScanNet200} and \textbf{SceneNN}, where it achieves competitive zero-shot instance segmentation performance, demonstrating reconstructive priors can effectively replace explicit depth supervision in online 3D perception.

Our contributions can be summarized as follows:
\begin{itemize}
    \item \textbf{A monocular, zero-shot 3D segmentation framework.} We present \method, the first system that performs online 3D instance segmentation directly from a monocular RGB stream by jointly leveraging reconstructive and visual foundation models.
    \item \textbf{Self-supervised query refinement and distillation.} We propose a spatial–semantic distillation strategy that enforces both instance-level discriminativeness and geometry-aware consistency without ground-truth annotations.
    \item \textbf{3D query index memory for temporal reasoning.} We design an index-based query memory mechanism that enables cross-frame association via spatial keys and contextual query retrieval.
    \item \textbf{Online mask fusion strategy.} We introduce a novel attention-based identity descriptor extracted from CUT3R’s state interactions, used to enhance mask fusion across frames along with the refined query descriptor.
\end{itemize}

\section{Related Works}
\label{sec:related}

\noindent\textbf{VFMs in 3D Segmentation}. Due to the scarcity of high-quality 3D annotations, adapting web-scale knowledge of 2D visual foundation models (VFMs)~\cite{oquab2023dinov2,kirillov2023sam,ravi2024sam2,zhai2023siglip,tschannen2025siglip2,simeoni2025dinov3} has become a promising solution to 3D scene segmentation~\cite{yang2023sam3d,lu2023corl,yin2024cvpr,yan2024cvpr,nguyen2024cvpr,nguyen2025any3dis,huang2024segment3d,boudjoghra2024iclr,takmaz2023nips,zhou2025ov3d,zhao2025cvpr}. 
For instance, SAM3D~\cite{yang2023sam3d} proposes to lift 2D segmentation masks from SAM~\cite{kirillov2023sam} to 3D point cloud and merge them upon geometric information. UnScene3D~\cite{rozenberszki2024cvpr} fuses 2D features from DINO~\cite{caron2021dino} with 3D features to create pseudo 3D masks for unsupervised self-training.  Any3DIS~\cite{nguyen2025any3dis} simplifies the complex mask merging process by applying 2D mask tracking based on SAM2~\cite{ravi2024sam2}. 
Despite their impressive performance, these methods work offline, {requiring the full RGB-D sequence to perform global optimization of mask grouping}.
To address the growing demands of embodied AI, many online methods~\cite{liu2022cvpr,mccormac2017semanticfusion,narita2019panopticfusion,zhang2020fusion,huang2021tog,xu2024memory} are proposed to process sequential RGB-D inputs. 
EmbodiedSAM~\cite{embodiedsam2025iclr} proposes the first VFM-assisted online framework and learns VFM mask lifting and merging in an online and fully supervised way {with ground truth for 3D instance masks}. 
Conversely, OnlineAnySeg~\cite{tang2025onlineanyseg} purely leverages 3D spatial information {using the ground truth geometry} to merge VFM masks without requiring 3D ground truth. In contrast to all previous methods, our method {works on monocular online zero-shot 3D segmentation without 3D ground truth geometry or segmentation masks.}

\noindent\textbf{Online RFMs}. While many traditional and learning-based methods address continuous online 3D geometry reconstruction, they typically face limitations such as requiring known camera intrinsics~\cite{engel2014eccv,forster2016tro,teed2021nips}, needing posed image inputs~\cite{kar2017nips,sun2021cvpr,sayed2022eccv}, or being restricted to object-centric scenarios~\cite{choy2016eccv,kar2017nips,yu2021cvpr}.
{Without the above limitations, }DUSt3R~\cite{wang2024dust3r} and subsequent works~\cite{leroy2024eccv,zhang2025monst3r,yang2025cvpr} demonstrate that accurate binocular reconstruction can be achieved by predicting pair-wise pointmaps in a feed-forward step, only taking uncalibrated and unposed images as input. 
Given their impressive generalizability and large-scale pretraining, these methods can be called reconstructive foundation models (RFMs). 
RFMs have since been extended to process varying numbers of views simultaneously~\cite{wang2025vggt,tang2025cvpr,keetha2025mapanything} or sequentially~\cite{wang20243dv,wang2025cut3r,wu2025point3r,liu2025slam3r,cabon2025cvpr,chen2025ttt3r}. 
We build our framework upon CUT3R~\cite{wang2025cut3r} to leverage its spatio-temporal priors learned from online 3D reconstruction. 
Unlike the explicit memory used in other online RFMs~\cite{wang20243dv,wu2025point3r,cabon2025cvpr}, CUT3R uniquely maintains a latent state as memory.
To leverage this non-interpretable representation, we design a novel state distribution token that extracts instance-level correspondences, achieving robust VFM mask matching.

\noindent\textbf{Monocular 3D Segmentation}. {Segmenting 3D objects based on monocular RGB input remains a significant challenge. One dominant category of approaches, based on NeRF~\cite{mildenhall2021nerf} or Gaussian Splatting (GS)~\cite{kerbl20233d}, lifts 2D segmentation masks into consistent 3D segmentations~\cite{zust2025panst3r,fu2022panoptic,zhu2024pcf,bhalgat2023contrastive,siddiqui2023panoptic,kundu2022panoptic,bingdm2023dmnerf,zhi2021place}. However, these methods typically require posed images and rely on offline, full-scene optimization. More recently, EA3D~\cite{zhou2025ea3d} develops a pose-free, monocular online GS framework, leveraging VFMs and an online RFM. However, its reliance on multi-step Gaussian optimization slows inference speed, and its evaluation, like other GS-based methods, focuses on 2D segmentation of novel views. Another line of work~\cite{wu2024panorecon,zhou2025eprecon} performs online panoptic reconstruction, simultaneously reconstructing and segmenting 3D scenes, but these methods are fully supervised on ground truth geometry and segmentation masks and still require posed monocular videos. In contrast to all previous methods, our method takes unposed monocular inputs, performs real-time online 3D segmentation, and is evaluated directly on its 3D segmentation results. }

\begin{figure*}[!t]
    \centering
    \includegraphics[width=\textwidth]{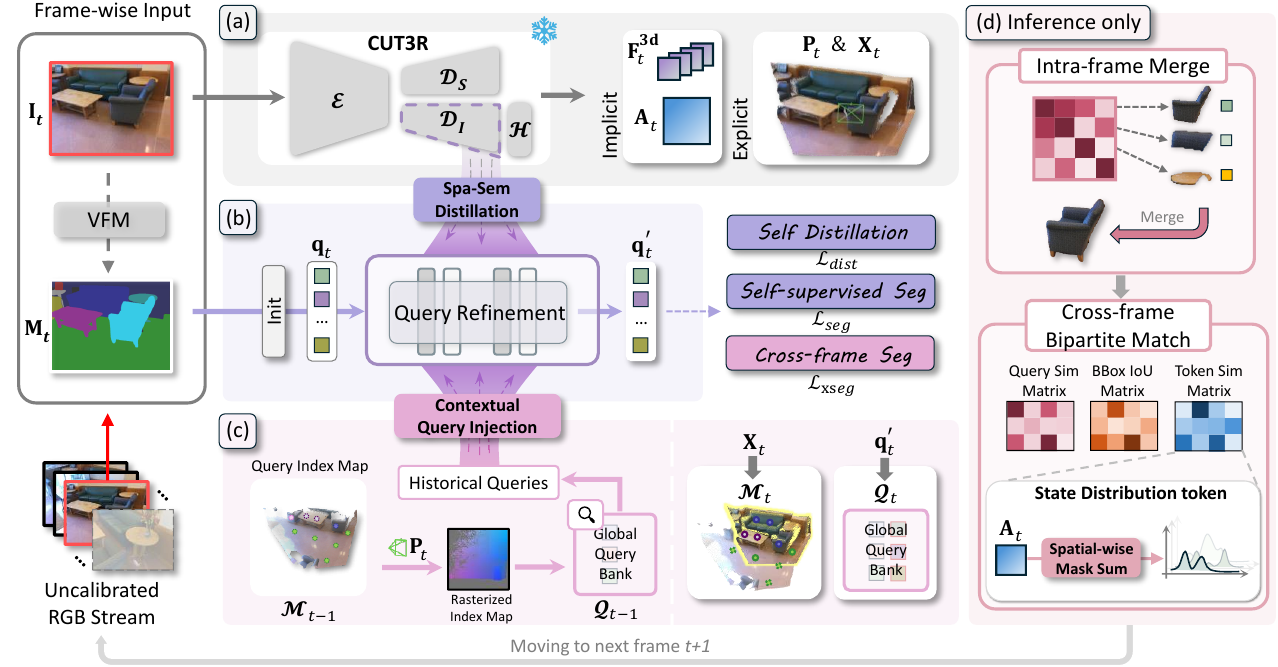}
    \caption{\textbf{Overview of MoonSeg3R}. The pipeline consists of four steps. (a) CUT3R takes an uncalibrated image $\textbf{I}_t$ as input to predict explicit geometry (pose $\textbf{P}_t$, world-coordinate pointmap $\textbf{X}_t$), and implicit representations (geometric features $\textbf{F}^{3d}_t$, state attention $\textbf{A}_t$). (b)  VFM masks $\textbf{M}_t$ are lifted and refined into 3D queries $\textbf{q}'_t$ through a transformer decoder, via spatial-semantic self-distillation supervision ($\mathcal{L}_{dist}$, $\mathcal{L}_{seg}$) (\cref{sec:feature_refinement}). (c) In parallel, we utilize $\textbf{P}_t$ to rasterize our explicit 3D query index memory $\mathcal{M}_{t-1}$, efficiently retrieving relevant historical queries from query bank $\mathcal{Q}_{t-1}$ for contextual query injection into query refinement process and cross-frame supervision ($\mathcal{L}_{xseg}$). The memory and bank are then updated using $\textbf{X}_t$ and $\textbf{q}'_t$, respectively (\cref{sec:memory}). (d) During inference, we first merge over-segmented instances and then perform bipartite matching, utilizing our novel state distribution token derived from state attention $\textbf{A}_t$ to enhance association robustness (\cref{sec:merge}).
    }
    \label{fig:overview}
    \vspace{-10px}
\end{figure*}

\section{Method}

Given a streaming monocular RGB sequence $\{\mathbf{I}_t\}_{t=1}^T$ with $\mathbf{I}_t \in \mathbb{R}^{H \times W \times 3}$, our goal is to incrementally reconstruct the scene geometry and segment object instances in an \emph{online zero-shot} manner, without access to ground-truth 3D geometry or instance masks. 
At each timestep $t$, \method predicts partial 3D instance masks by unprojecting 2D masks from a Visual Foundation Model (VFM) and incrementally associates them across frames to form consistent 3D instances over time.

\noindent\textbf{Preliminaries: CUT3R.} \method utilizes the geometric prior in a pre-trained online 3D reconstruction model, CUT3R~\cite{wang2025cut3r}, to perform zero-shot online segmentation. CUT3R recurrently processes a stream of RGB images, predicting the \textit{explicit} geometry information (depth, pose, pointmap) of all frames. At each timestep, the encoded features of the input image interact (through cross attention) with a persistent ``state token'' that serves as a compressed latent representation of the entire scene, reading from the state and updating it. The updated image encodings are then used to decode the geometric outputs. The state tokens in CUT3R have been found to \textit{implicitly} encode the scene geometry, enabling ``reading out'' the geometry of unseen regions during reconstruction. We exploit both the implicit and explicit geometric information in CUT3R for accurate 3D segmentation.

\subsection{Online Segmentation Framework}
\label{sec:framework}

At each timestep $t$, the incoming image $\mathbf{I}_t$ is processed by two pretrained foundation models:
\begin{equation}
\begin{aligned}
(\mathbf{X}_t, \mathbf{P}_t, \mathbf{F}_t^{3d}, \mathbf{A}_t) &= \text{RFM}(\mathbf{I}_t),\\
\mathbf{M}_t &= \text{VFM}(\mathbf{I}_t),
\end{aligned}
\end{equation}
where $\mathbf{X}_t$ denotes the predicted world-coordinate pointmap, $\mathbf{P}_t$ the estimated camera pose, $\mathbf{F}_t^{3d}$ the 3D geometric feature map, and $\mathbf{A}_t$ the state attention weights. 
Each 2D instance mask $\mathbf{M}^i_t$ is unprojected into a partial 3D mask using $\mathbf{X}_t$. 

In our experiments we choose CUT3R~\cite{wang2025cut3r} as the RFM providing online 3D reconstruction and CropFormer~\cite{qi2023iccv} as the VFM, providing per-frame segmentation masks. 
The state-branch attention maps $\mathbf{A}_t$ indicate how the state tokens within CUT3R attend to image patches and later serve as temporal cues for instance correspondence (Sec.~\ref{sec:merge}).

\paragraph{3D Prototype Representation.}
For a single frame $\mathbf{I}_t$, we represent each object as a {3D prototype} $\mathbf{q}^i_t \in \mathbb{R}^d$ by lifting the corresponding 2D instance mask $\mathbf{M}^i_t$ into $\mathbf{q}^i_t$ via masked average pooling over concatenated geometric and semantic features:
\begin{equation}
\label{eq:prototype}
\mathbf{F}_t = [\mathbf{F}^{3d}_t, \mathbf{F}^{2d}_t], \quad
\mathbf{q}^i_t = \eta \!\left(\frac{\sum_{u,v} \mathbf{F}_t(u,v)\mathbf{M}^i_t(u,v)}{\sum_{u,v}\mathbf{M}^i_t(u,v)}\right),
\end{equation}
where $u,v$ are the pixel coordinates and $\eta(\cdot)$ is a learned linear projection.
As $\mathbf{F}^{3d}_t$ mostly captures the geometric information, we augment them with $\mathbf{F}^{2d}_t$, 2D semantic features from DINOv3~\cite{simeoni2025dinov3}.

The key challenge of 3D instance segmentation in an online setting is to associate prototypes with the correct object instances accurately not only at the current step but also across the previous timesteps.
In other words, our setting requires to estimate accurate segmentation masks even when the VFM provides over or under-segmented proposals, to associate the prototypes of the same object appearing across multiple frames and merge them, and to recognize when a new object instance appeared that is not to be merged.
All together, this demands the prototypes to encode the correct spatial, semantic and chronological information in an accurate and efficient manner.

Motivated by these challenges, our model contains (i) query refinement module (see~\cref{sec:query_refinement}) to associate them with the required spatial, semantic and chronological information, (ii) spatial-semantic distillation module (see~\cref{sec:feature_refinement}) to provide the self-supervision signal for training the query refinement module, (iii) contextual memory module (see~\cref{sec:memory}) to store the previously seen queries and to associate the current queries with the previous ones, and finally (iv) an online mask fusion module (see~\cref{sec:merge}) to merge queries from previously seen partial observations and obtain the 3D segmentation predictions.

\subsection{Query Refinement}
\label{sec:query_refinement}

We first use a query decoder $\phi$, a lightweight feed-forward projector, to refine $\mathbf{q}_t$ in \cref{eq:prototype} through masked cross-attention:
\begin{equation}
\mathbf{q}'_t \leftarrow \phi\!\left(\text{Attn}\!\left(\mathbf{q}_t,\, \mathbf{F}_t,\, \mathbf{M}_t\right)\right).
\end{equation}
This refinement step further adapts the semantic and geometric features to the target segmentation mask.
As the refined query only contains the information from the current timestep $t$, we further incorporate contextual information from previously seen queries $\mathcal{Q}^{\text{ctx}}_t$, which we describe in \cref{sec:memory}, through an additional cross-attention layer:
\begin{equation}
\label{eq:contextualrefinement}
\mathbf{q}'_t \leftarrow \phi(\text{Attn}(\mathbf{q}'_t,\, \mathcal{Q}^{\text{ctx}}_t)).
\end{equation}

\subsection{Spatial-Semantic Distillation}
\label{sec:feature_refinement}
As we focus on the unsupervised setting, with no access to the ground-truth segmentation masks during training, we propose a self-supervised strategy to train the parameters of $\eta$ and $\phi$.
We would like each refined query $\mathbf{q}'_t$ to encode the sufficient information to recover the original mask $\mathbf{M}_t$ in the pixel space.
One potential solution is to first project the feature map $\mathbf{F}_t$ to lower dimensionality through a MLP $\psi$, and then apply the refined query pointwisely to the projected feature map to obtain the corresponding mask:
\begin{equation} 
\label{eq:segloss}
\mathcal{L}_{\text{seg}} = \text{BCE}(\sigma(\psi(\mathbf{F}_t)\odot\mathbf{q}'_t),\mathbf{M}) 
\end{equation} where $\sigma$ is sigmoid function.
While this formulation encourages learning queries to discriminate the corresponding instance against the remaining ones in the same frame, it does not necessarily enforce the queries to retain both geometric and spatial information.
In our preliminary experiments, we observed that geometry information from $\mathbf{F}^{\text{3d}}$ was discarded.
We argue that $\mathbf{F}^{\text{2d}}$ are sufficiently rich for satisfying the training objective.
On the other hand, this does not facilitate cross-frame association of queries.

To address this, we add an additional objective to preserve the internal structures of the concatenated $\mathbf{F}$ fusing $\mathbf{F}^{2d}$ and $\mathbf{F}^{3d}$. 
Specifically, we utilize Gram matrix as it computes the pairwise dot product of patch features, and provides patch-level structural guidance, while allowing local features to be freely updated.
We compute the Gram matrix of the reference features $\mathbf{F}$ as
\[ \mathbf{G}=\frac{\psi(\mathbf{F})\psi(\mathbf{F})^\top}{\|\psi(\mathbf{F})\|\|\psi(\mathbf{F})\|},
\]and similarly compute $\mathbf{G}^{2d}$ from $\mathbf{F}^{2d}$ for 2D semantic guidance and $\mathbf{G}^{3d}$ from $\mathbf{F}^{3d}$ as 3D geometric guidance. 
We then design a Gram distillation loss that encourages the reference features to preserve essential patterns from CUT3R and DINO, while simultaneously enhancing their discriminativeness. 
\begin{equation}
\label{eq:distloss}
    \mathcal{L}_{\text{dist}} = \|\mathbf{G}-\mathbf{G}^{2d}\|^2_F + \|\mathbf{G}-\mathbf{G}^{3d}\|^2_F 
\end{equation}

\subsection{Contextual Query Indices as Memory}
\label{sec:memory}

The refined queries $\mathbf{q}'_t$ encode the spatial–semantic information of each observed instance in the current view. 
However, monocular inputs provide only partial information about each object, leading to incomplete or fragmented 3D representations.
To ensure temporal consistency and leverage historical context, we introduce a \textbf{3D Query Index Memory (QIM)}, a lightweight, index-based memory mechanism that links current queries to relevant historical counterparts across time.

\noindent\textbf{Memory Representation.}
At each timestep $t$, we maintain two complementary memory structures:
\begin{itemize}
    \item a \emph{global query bank} $\mathcal{Q}_{t-1} \in \mathbb{R}^{n^q_{\text{total}} \times d}$ that stores all historical query features up to time $t\!-\!1$, where $n^q_{\text{total}}$ denotes the total number of queries and $d$ their feature dimension;
    \item a \emph{query index map} $\mathcal{M}_{t-1} \in \{0,1\}^{n^k_{\text{total}} \times n^q_{\text{total}}}$ that records which queries were associated with each 3D spatial key (described below) in the scene.
\end{itemize}

\noindent\textbf{Memory Update.}
Given the predicted pointmap $\mathbf{X}_t$, the VFM-generated masks $\mathbf{M}_t$, and the corresponding refined queries $\mathbf{q}'_t$, 
we first append $\mathbf{q}'_t$ to the global query bank:
\begin{equation}
\mathcal{Q}_t = [\,\mathcal{Q}_{t-1};\, \mathbf{q}'_t\,].
\end{equation}
To efficiently map queries to spatial locations, we sample a sparse set of $n^k_t$ 3D \emph{spatial keys} $\{\mathbf{k}^i_t\}_{i=1}^{n^k_t}$ from $\mathbf{X}_t$ by { average pooling on world-coordinate pointmaps}. 
Each key represents a compact local region of reconstructed geometry.

Next, we downsample the segmentation masks $\mathbf{M}_t$ to the same spatial resolution as $\{\mathbf{k}^i_t\}$ and establish associations between each query and its corresponding spatial key. 
For key $\mathbf{k}^i_t$ and query $\mathbf{q}'^j_t$, we define a binary association:
\begin{equation}
\mathcal{M}_t(i,j) = 
\begin{cases}
1, & \text{if } \mathbf{k}^i_t \text{ lies within mask } \mathbf{M}^j_t,\\
0, & \text{otherwise.}
\end{cases}
\end{equation}

\noindent\textbf{Contextual Query Retrieval.}
During the processing of frame $t$, we exploit CUT3R’s predicted camera pose $\mathbf{P}_t$ to project all stored spatial keys into the current camera coordinate system.
Rasterizing these projected keys forms an \emph{index map} $\mathbf{I}^{\text{ind}}_t \in \{0,1\}^{H \times W \times n_q}$, where each pixel contains binary indicators of visible historical queries. 
The corresponding historical queries are then retrieved from the global bank:
\begin{equation}
\mathcal{Q}^{\text{ctx}}_t = \text{Retrieve}(\mathbf{I}^{\text{ind}}_t,\, \mathcal{Q}_{t-1}).
\end{equation}
These contextual queries are injected into the refinement process in \cref{eq:contextualrefinement}, allowing each current query to attend to semantically and geometrically similar queries from previous frames.

\noindent\textbf{Cross-Frame Supervision}. 
To ensure that our model learns to encode the chronological context into the refined queries, we introduce a cross-frame objective that compels the updated queries to attend to the retrieved historical information. Given the rasterized index map $\mathbf{I}^{ind}_t$, the retrieved historical queries at each spatial location can be averaged, therefore forming a contextual feature map $\mathbf{F}^{ctx}_t\in \mathbb{R}^{H\times W\times d}$. 
We can therefore calculate the cross-frame version of $\mathcal{L}_{\text{seg}}$ as:
\begin{equation}
    \label{eq:xsegloss}
    \mathcal{L}_{\text{xseg}} = \textnormal{BCE}(\sigma(\mathbf{F}_t^{ctx}\odot \mathbf{q}'_t) \odot \mathbf{M}^{ind}, \mathbf{M}_t \odot \mathbf{M}^{ind})
\end{equation}
where $\mathbf{M}^{ind}$ is used to mask out the loss values in the areas in which no historical information is rasterized in the current view.

\noindent\textbf{Training Objective}.
Finally, we minimize the weighted sum of \cref{eq:segloss}, \cref{eq:distloss} and \cref{eq:xsegloss} with respect to the network parameters:
\[
\min_{\eta,\phi,\psi} \lambda_{\text{seg}} \mathcal{L}_{\text{seg}}  + \lambda_{\text{dist}}\mathcal{L}_{\text{dist}} + \lambda_{\text{xseg}}\mathcal{L}_{\text{xseg}},
\]
where $\lambda_{\text{seg}},\lambda_{\text{dist}},\lambda_{\text{xseg}}$ are the scalar loss weights. 

\begin{table*}[!ht]
    \centering
\renewcommand{\arraystretch}{1.2}
\small

   {
    \begin{tabular}{l|ccc|ccc|ccc|c}
    \specialrule{1.0pt}{0pt}{0pt}
          \multirow{2}{*}{\textbf{Method}} & \multirow{2}{*}{\textbf{Online}} &\multirow{2}{*}{\textbf{Zero-Shot}} & \multirow{2}{*}{\textbf{Input}} & \multicolumn{3}{c|}{\textbf{ScanNet 200}}& \multicolumn{3}{c|}{\textbf{SceneNN}} &  \textbf{Speed}\\ 
            & & & &  $AP$ & $AP_{50}$ & $AP_{25}$  & $AP$ & $AP_{50}$ & $AP_{25}$  & ms\\
    \specialrule{1.0pt}{0pt}{0pt}
            \multicolumn{11}{c}{Posed RGB-D}\\
        \midrule
        \textcolor{gray}{EmbodiedSAM}~\cite{embodiedsam2025iclr}
             & \textcolor{gray}{\ding{51}} &   \textcolor{gray}{\ding{55}} & \textcolor{gray}{\faImages \ + \faCamera} &  \textcolor{gray}{28.8} &   \textcolor{gray}{42.7} & \textcolor{gray}{54.2} & \textcolor{gray}{20.1}& \textcolor{gray}{32.5}& \textcolor{gray}{46.3}&\textcolor{gray}{80}\\
        \cmidrule{1-11}
        \textcolor{gray}{OVIR-3D}~\cite{lu2023corl} & \textcolor{gray}{\ding{55}} & \textcolor{gray}{\ding{51}} & \textcolor{gray}{\faImages \ + \faCamera}& \textcolor{gray}{14.4} & \textcolor{gray}{27.5} & \textcolor{gray}{38.8} & \textcolor{gray}{12.3}& \textcolor{gray}{24.4}& \textcolor{gray}{34.6}& \textcolor{gray}{--}\\
         \textcolor{gray}{MaskClustering}~\cite{yan2024cvpr} &\textcolor{gray}{\ding{55}} & \textcolor{gray}{\ding{51}} & \textcolor{gray}{\faImages \ + \faCamera} & \textcolor{gray}{19.7} & \textcolor{gray}{36.4} & \textcolor{gray}{51.4} & \textcolor{gray}{16.3}& \textcolor{gray}{31.7}& \textcolor{gray}{46.2}&\textcolor{gray}{--}\\
         \textcolor{gray}{SAM3D}~\cite{yang2023sam3d} & \textcolor{gray}{\ding{51}} & \textcolor{gray}{\ding{51}} & \textcolor{gray}{\faImages \ + \faCamera} & \textcolor{gray}{9.6} & \textcolor{gray}{24.8} & \textcolor{gray}{49.6} & \textcolor{gray}{9.1} & \textcolor{gray}{21.3} & \textcolor{gray}{43.4}& \textcolor{gray}{125}\\
        \textcolor{gray}{OnlineAnySeg}~\cite{tang2025onlineanyseg} & \textcolor{gray}{\ding{51}} & \textcolor{gray}{\ding{51}} & \textcolor{gray}{\faImages \ + \faCamera} & \textcolor{gray}{18.6} & \textcolor{gray}{36.1} & \textcolor{gray}{53.5} & \textcolor{gray}{18.1}& \textcolor{gray}{35.3}& \textcolor{gray}{59.5}&\textcolor{gray}{3000*}\\
        \midrule
        \multicolumn{11}{c}{Monocular}\\
        \midrule
         OnlineAnySeg-M & \ding{51} & \ding{51} & \faImage & 13.4 & 26.8 & 43.2 &13.2 & 28.7& 51.2&3000 + \underline{66}\\
         \textbf{\method(Ours)} & \ding{51} & \ding{51} & {\faImage} & 16.7& 33.3 & 50.0 & 14.3& 31.4& 48.4&55 + \underline{66}\\ 
         
\specialrule{1.0pt}{0pt}{0pt}
    \end{tabular}
    }
    \caption{{\bf Results on ScanNet 200 and SceneNN.}  While previous methods require posed RGB-D inputs ($\text{\faImages} + \text{\faCamera}$), our method takes monocular images ($\text{\faImage}$) and performs simultaneous reconstruction and segmentation. Regarding inference speed, we report the time cost for mask fusion, and additionally report the running time of RFM for geometry reconstruction as the underlined numbers in the monocular setting. MoonSeg3R clearly outperforms OnlineAnySeg-M, which is a monocular version of OnlineAnySeg taking predicted depths and poses as input, while being significantly faster, and comes close in performance to methods with posed RGB-D input while using monocular RGB inputs. *: Since OnlineAnySeg reports the time cost of online fusion stage~\cite{lan2025boxfusion}, we measure and report the speed of its full algorithm. 
    }
\label{tab:results}
\vspace{-10px}
\end{table*}

\subsection{Inference-time Online Mask Fusion}
\label{sec:merge}

{In inference time, different from fully-supervised methods~\cite{embodiedsam2025iclr,wang2025nips} that learn mask merging in different frames with ground truth 3D instance masks, we determine whether partial 3D masks $\mathbf{M}_t^{3d}$ should be merged based on multiple mask representations: query feature, bounding box and a novel state distribution token. To obtain the bounding box of each mask, we find its 3D spatial keys from $\mathcal{M}_{t-1}$, which correspond to the 3D point coordinates of the mask, and simply find the max and min coordinates to calculate the bounding box.} 

\noindent\textbf{State Distribution Token}. During the image-state interaction process of CUT3R, the state-branch cross-attention $\mathbf{A}\in \mathbb{R}^{n_s\times (h\times w)}$ indicates the attention weights allocated by each of the $n_s$ state tokens to the $h\times w$ image patches. Our finding is that this attention distribution provides a unique and temporally stable identity for segmented instances across neighboring frames. We call this representation state distribution token. As illustrated in the right bottom of Fig.~\ref{fig:overview}, the state distribution token of an instance can be obtained by applying its 2D instance mask $\mathbf{M}_t$ over the spatial dimension of $\mathbf{A}$ and performing a masked summation to aggregate the total attention allocated by each state token to the instance area: 

\begin{equation}
    \mathbf{s}^i_t = \sum_{j=1}^{h \times w} (\mathbf{A}_t \odot \mathbf{M}^i_t) \in \mathbb{R}^{n_s}
\end{equation}

\noindent\textit{Intra-Frame Merge}. As VFM usually oversegments an instance into multiple parts, we first merge the oversegmented parts within the current frame, leveraging the discriminative queries $\mathbf{q}'_t$ obtained previously. The pairwise query similarity matrix is calculated as $\mathbf{E}^{intra}_t = \frac{\langle\mathbf{q}'_{t},\mathbf{q}'_t\rangle}{{\|\mathbf{q}'_t\|\|\mathbf{q}'_t\|}} \in \mathbb{R}^{n_q\times n_q}$, with the diagonal elements equal to 1 representing self-similarity. We merge any mask pair $(\mathbf{M}^i_t, \mathbf{M}^j_t)$ whose similarity $\mathbf{E}_t^{intra}{(i,j)}$ exceeds a predefined threshold.

\noindent\textit{Cross-Frame Bipartite Match}. After intra-frame merging, we calculate the similarity matrices between current-frame masks and the existing ones as 
\begin{equation} \mathbf{E}^{ij} = \frac{\langle\mathbf{q}^{prev, i}, \mathbf{q'}^{j}_t\rangle}{\|\mathbf{q}^{prev, i}\|\|\mathbf{q'}^{j}_t\|} + \frac{\langle\mathbf{s}^{prev, i},\mathbf{s}_{t}^j\rangle}{\|\mathbf{s}^{prev,i}\|\|\mathbf{s}_{t}^j\|} + {\textnormal{IoU}(\mathbf{b}^{prev,i}, \mathbf{b}_t^j)} \end{equation}
where the three items respectively indicate the query similarity, state distribution token similarity and bounding box IoU between previous masks and the new masks. Similar to~\cite{embodiedsam2025iclr}, we prune $\mathbf{E}$ by setting scores below a predefined threshold to $-\infty$, and use $-\mathbf{E}$ as the cost to assign each new mask to existing masks. If a new mask fails to find a match, we register it as a new instance.

\noindent\textit{Mask Update}. When two masks are merged, we update their query features and state distribution tokens by averaging, and compute the union of their spatial keys to update the mask position information.

\begin{figure*}
    \centering
    \includegraphics[width=0.9\textwidth]{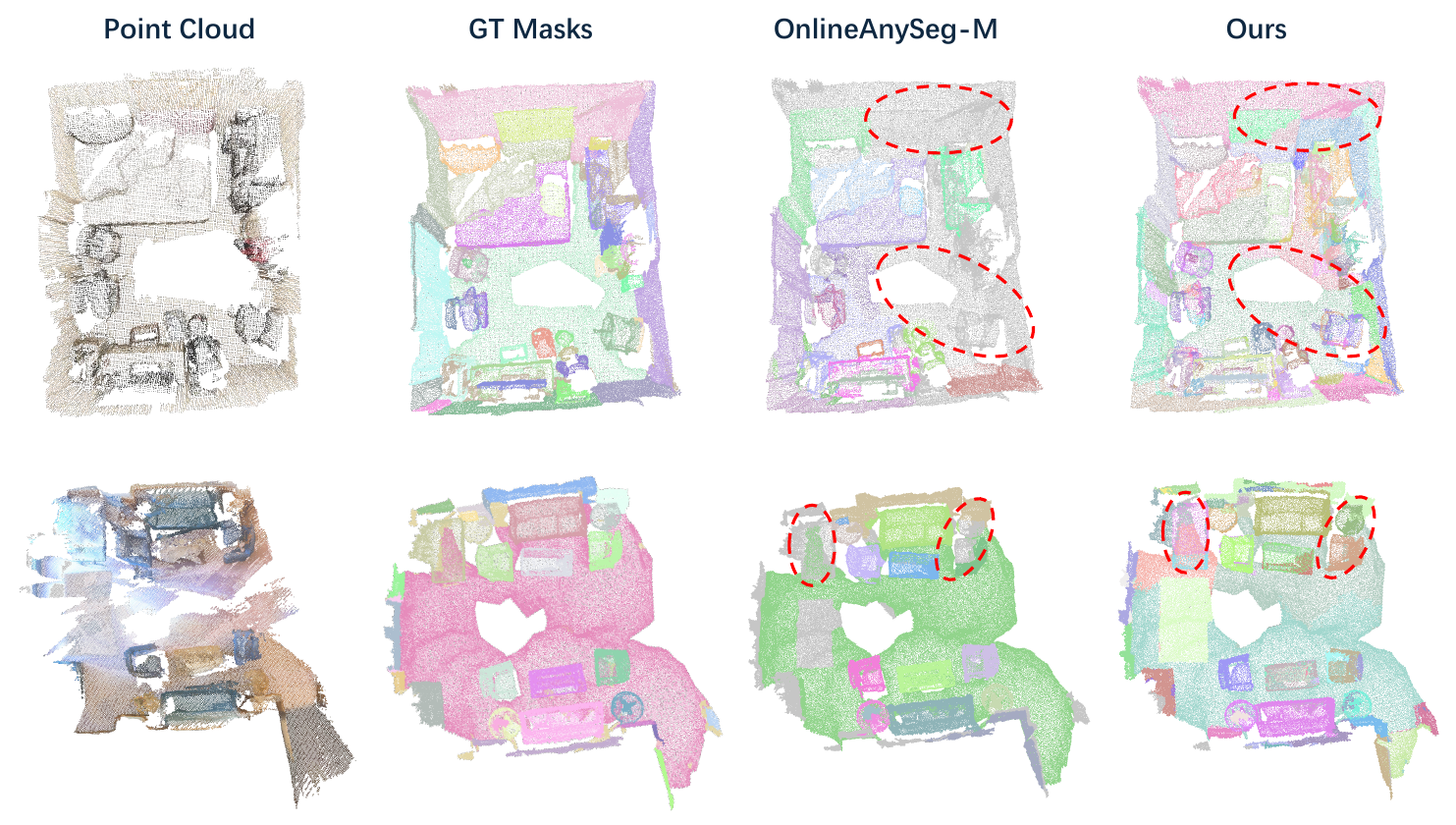}
    \caption{\textbf{Qualitative Comparison.} Qualitative examples of OnlineAnySeg-M and our method on ScanNet200 sequences. These results visually demonstrate that \method achieves superior instance segmentation. OnlineAnySeg-M, in contrast, tends to fail in associating masks, which leaves significant unsegmented areas, as shown in the red dashed circles. The segmentation results are unprojected to ground truth point cloud for visualization. }
    \label{fig:results}
    \vspace{-10px}
\end{figure*}

\section{Experiments}

\subsection{Implementation Details}
\label{sec:implementation}

The query refinement network $\phi$ is implemented as 3 transformer-based decoder layers, each consisting of two consecutive cross-attention layers, one self-attention and feed-forward layer. The query projection layer $\eta$ and the feature refinement network $\psi$ are both implemented as a series of MLP layers. To train MoonSeg3R, we randomly sample 16 adjacent RGB frames of size 512$\times$384 from each scene in ScanNet, and generate their VFM masks using FastSAM. CUT3R and DINOv3 are frozen during training. We train MoonSeg3R for 100 epochs, with AdamW as our optimizer and a learning rate of 1e-4 (with a cosine decay to 1e-5). The loss weights $\lambda_{seg}$, $\lambda_{xseg}$ and $\lambda_{dist}$ are respectively set to 1, 0.5 and 0.1. The training takes 6 hours on 4 NVIDIA RTX A6000 GPUs, with a batch size of 4 on each. During testing, for fair comparison with~\cite{tang2025onlineanyseg}, we generate VFM masks using CropFormer~\cite{qi2023iccv}, and set keyframe sampling interval as 10 on ScanNet200 and 20 on SceneNN. The thresholds for intra-frame merging and cross-frame matching are respectively 0.8 and 1.8. We test our method on an NVIDIA RTX A6000 GPU.

\subsection{Experimental Settings}

\noindent\textbf{Dataset}. We conduct experiments on 3D instance segmentation on two real-world benchmarks, ScanNet200~\cite{dai2017cvpr,rozenberszki2022eccv} and SceneNN~\cite{hua2016scenenn}. ScanNet200 is an indoor dataset comprising 1513 room-level sequences, each annotated with instance-level segmentation and labels across 200 categories. Consistent with the compared methods, we evaluate our approach on the
 validation set, which includes 312 scenes. SceneNN contains 50 high-quality scanned scenes, in which we select 12 clean sequences for testing following previous works~\cite{tang2025onlineanyseg,embodiedsam2025iclr}.  

\noindent\textbf{Evaluation Metric}. We adopt Average Precision (AP) as the metric. We follow~\cite{yan2024cvpr,tang2025onlineanyseg} to report the results under IoU thresholds of 25\% and 50\%, and the mean AP across thresholds from 50\% to 95\%, denoted as $AP_{25}$, $AP_{50}$, and $AP$, respectively.

\subsection{Experimental Results}

We compare with VFM-assisted RGB-D based 3D segmentation methods on ScanNet200 and SceneNN, including offline zero-shot methods OVIR-3D~\cite{lu2023corl}, MaskClustering~\cite{yan2024cvpr}, online fully-supervised method~\cite{embodiedsam2025iclr} and zero-shot methods~\cite{yang2023sam3d,tang2025onlineanyseg}. To compare with these methods, we unproject the merged masks to the ground truth point cloud during testing, so as to calculate the average precision \emph{w.r.t.} the ground truth. The results are copied from the reported performance in~\cite{tang2025onlineanyseg}. 
Additionally, since \method is the only method capable of \emph{monocular} online zero-shot segmentation, we build a monocular baseline by inputting CUT3R predicted poses and depths to OnlineAnySeg, which is denoted by OnlineAnySeg-M. Class-agnostic instance segmentation results are reported in Tab.~\ref{tab:results}.

\noindent\textbf{Online 3D Segmentation}. Taking only RGB images as input, our method outperforms OVIR-3D and SAM3D, which are respectively RGB-D based offline and online methods, on both benchmarks. {Though online methods EmbodiedSAM~\cite{embodiedsam2025iclr} and OnlineAnySeg~\cite{tang2025onlineanyseg} achieve higher results, their performance depends heavily on ground truth geometry. For instance, replacing ground truth pose and depths with the predictions from CUT3R results in a significant performance drop on OnlineAnySeg-M, demonstrating that existing methods designed for posed RGB-D sequences struggle to generalize to the monocular setting. On the other hand, the proposed MoonSeg3R can robustly associate and merge different 2D masks without requiring external geometry or 3D segmentation supervision, surpassing OnlineAnySeg-M by +3.3 on ScanNet200 and 1.1 on SceneNN regarding AP.}

\noindent\textbf{Speed Analysis}. As shown in the rightmost column of Tab.~\ref{tab:results}, MoonSeg3R achieves the fastest mask fusion speed and ranks second overall, while simultaneously performing reconstruction and segmentation. In contrast to the state-of-the-art zero-shot method OnlineAnySeg, which requires time-consuming per-instance CLIP feature extraction and iterative mask graph traversal, our method  infers in one feed-forward step, benefiting from a lightweight decoder, sparse memory, and efficient point rasterization.

\noindent\textit{Qualitative Results}. In Fig.~\ref{fig:results}, we provide qualitative comparison between OnlineAnySeg-M and our method.

\begin{figure}
    \centering
    \includegraphics[width=\columnwidth]{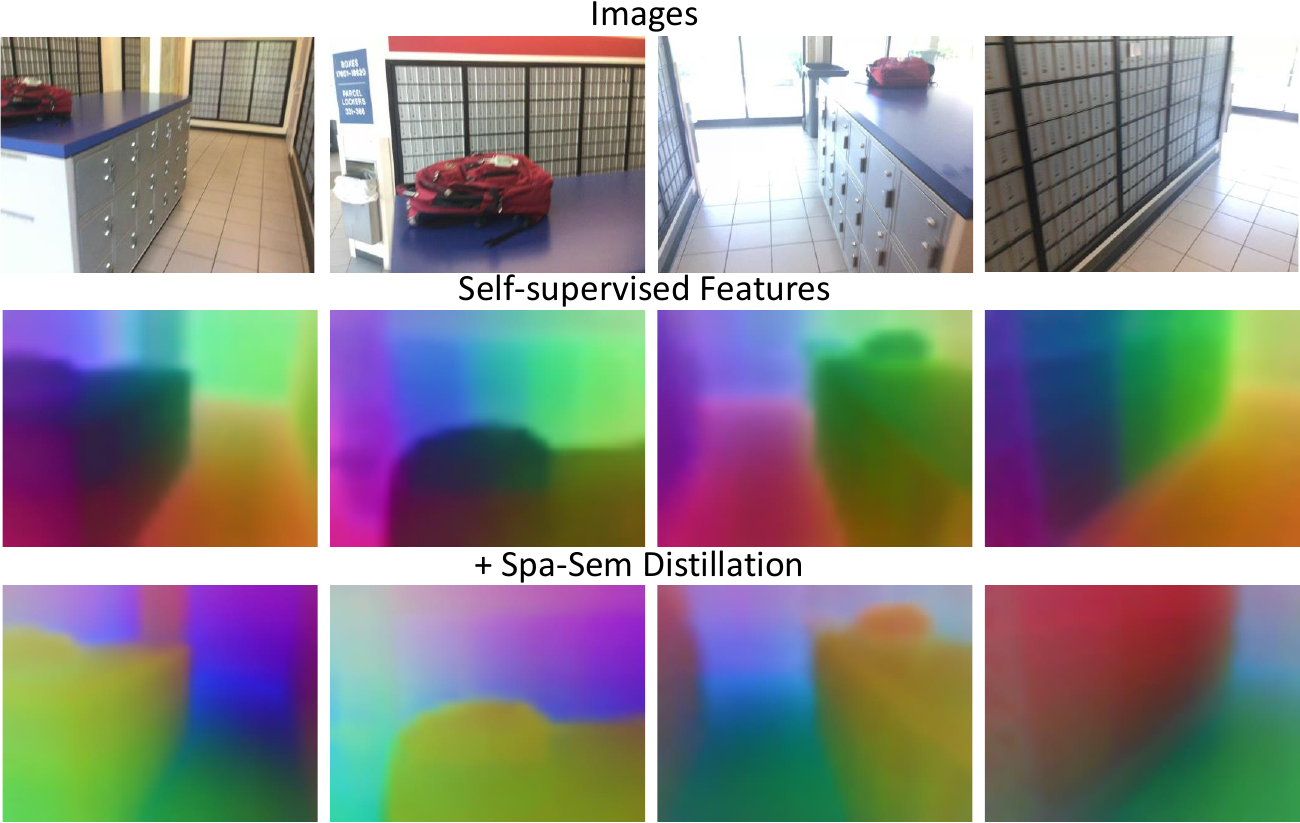}
    \caption{\textbf{Distilled Feature Visualization.} Top row: The original images. Middle row: Reference features trained only with self-supervision. These features show a fixed spatial pattern (purple to yellow) that is not correlated with the actual spatial location. Bottom row: Features trained with spatial-semantic distillation. This strategy mitigates the feature degradation by preserving essential structural patterns from the foundation models. The resulting features are object-aware, as the same bag area remains consistent across views, while the features of different cabinets properly reflect the 3D spatial variation.
    }
    \label{fig:results_feat}
    \vspace{-5px}
\end{figure}

\begin{table}[!t]
    \centering


    {
    \begin{tabular}{l|ccc}
    \specialrule{1.0pt}{0pt}{0pt}
    {\textbf{Method}} & $AP$ & $AP_{50}$ & $AP_{25}$\\
    \midrule
    $\mathbf{F}^{2d}$ & 7.2 & 19.6 & 43.7\\
    $\mathbf{F}^{3d}$ & 6.4 & 17.5 & 41.1\\
    $\mathbf{F}^{2d} + \mathbf{F}^{3d}$ & 8.1 & 19.7 & 42.3\\
    \midrule
    + Query Refinement & 12.5 & 27.7 & 46.6\\
    + SSD & 13.5 & 29.3 & 48.5\\
    + QIM & 15.9 & 32.8 & 49.4\\
    + SDT (Ours) & \textbf{16.7} & \textbf{33.3} & \textbf{50.0}\\
         
\specialrule{1.0pt}{0pt}{0pt}
    \end{tabular}
    }
    \caption{\textbf{Ablation Study.} The individual contributions of the proposed components in MoonSeg3R on ScanNet200.  
    }
    \vspace{-15px}
\label{tab:ablation}

\end{table}

\subsection{Ablation Study}

In Tab.~\ref{tab:ablation}, we show the ablated results to validate the effectiveness of each proposed component. We use CropFormer to generate VFM masks during testing and report the results on ScanNet200.

\noindent\textbf{Baseline}. We first establish a baseline to evaluate the raw power of the foundation model features. This baseline computes feature similarity matrices using mask-averaged 2D DINO features ($\mathbf{F}^{2d}$) and 3D geometric CUT3R features ($\mathbf{F}^{3d}$) when associating masks across timesteps. As shown in the first three rows of Tab.~\ref{tab:ablation}, using either $\mathbf{F}^{2d}$ or $\mathbf{F}^{3d}$ alone yields poor performance as the former lacks 3D grounding while the latter lacks semantic awareness. Combining $\mathbf{F}^{2d}$ and $\mathbf{F}^{3d}$ brings marginal improvement, indicating the original features are inadequate to associate same-instance masks while separating different ones.

\noindent\textbf{Query Refinement}. We then add our query refinement network $\phi$ and apply $\mathcal{L}_{seg}$ to learn to lift masks into queries in a self-supervised manner. The performance is largely boosted by +4.4 on $AP$ and +8.0 on $AP_{50}$, demonstrating that the self-supervised query learning strategy successfully generates a discriminative query prototype for each instance.

\noindent\textbf{Spatial-Semantic Distillation}. To avoid the feature degradation during feature refinement, we further distill the internal structural information from foundation models to guide this process, which is denoted by +SSD in Tab.~\ref{tab:ablation}. The distillation strategy brings an improvement of +1.0 on AP, as it allows the feature refining to preserve the rich structure information from foundation models, as well as freely updating local feature information. 

\noindent\textit{Reference Feature Visualization}. In Fig.~\ref{fig:results_feat}, we visually validate that, while the self-supervised query learning contributes to higher object discriminativeness in reference features, the features learn a shortcut with fixed spatial pattern. By applying spatial-semantic distillation strategy, this problem is mitigated.

\noindent\textbf{Query Index Memory}. Incorporating the query index memory and the corresponding cross-frame segmentation loss $\mathcal{L}_{xseg}$ (+QIM in Tab.~\ref{tab:ablation}) further improves the AP by +2.4, demonstrating that the contextual information efficiently supplements the limited information from a single-view observation.

\begin{figure}
    \centering
    \includegraphics[width=\linewidth]{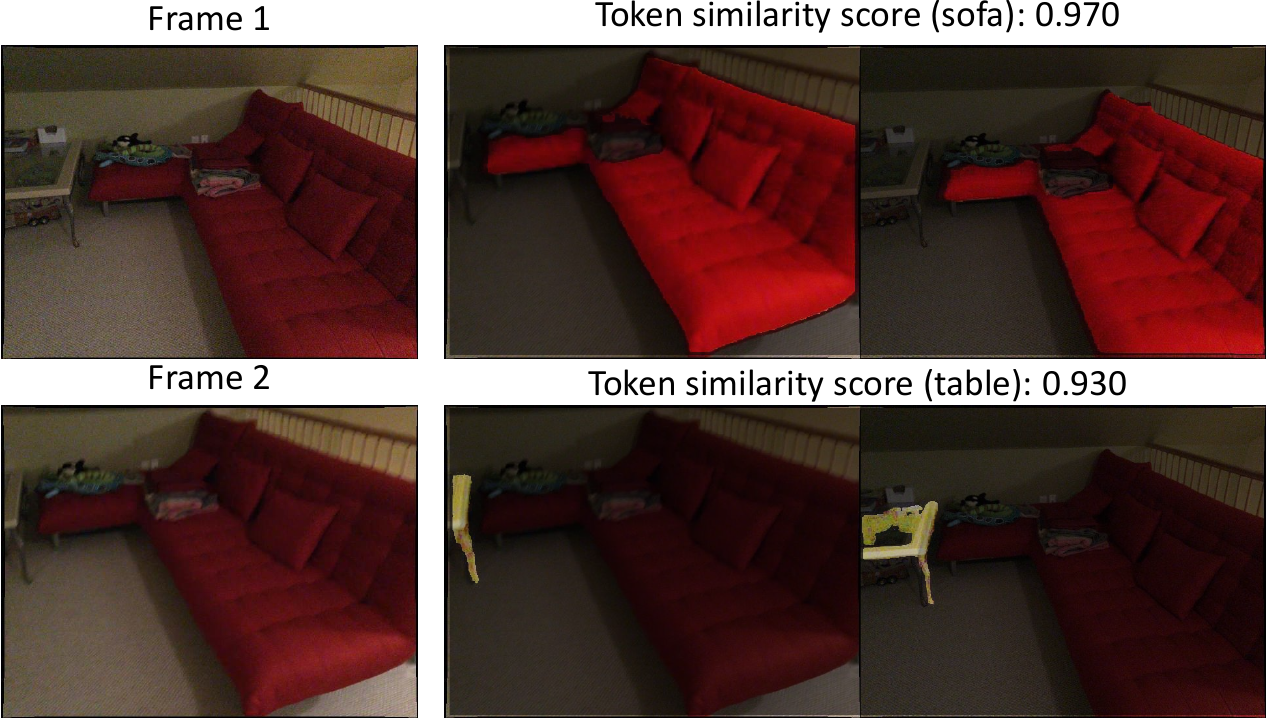}
    \caption{\textbf{State Distribution Similarity.} For two consecutive frames, we extract the state distribution tokens for all instances and compute their cross-frame pairwise similarities. Tokens belonging to the same instances always exhibit the highest similarity scores, both for large, fully-visible objects (sofa) and small, partially observed objects (table).}
    \label{fig:state_vis}
    \vspace{-5px}
\end{figure}

\noindent\textbf{State Distribution Token}. Finally, we integrate the state distribution token at inference time to assist bipartite matching, denoted as "+SDT" in Tab.~\ref{tab:ablation}. This component improves the AP by +0.8. As visualized in Fig.~\ref{fig:state_vis}, the SDT robustly matches instances across frames, whether they are large, fully-visible objects or small, partially observed ones.

\section{Conclusion}

This paper introduces MoonSeg3R, the first framework to perform monocular, online, and zero-shot 3D instance segmentation without requiring ground-truth geometry from posed RGB-D sequences. 
MoonSeg3R builds on the reconstructive and semantic priors by transforming VFM masks into discriminative 3D queries with self-supervised refinement and distillation, and incorporating cross-view contextual information through a query index memory. At last, a novel state distribution token is extracted from CUT3R's state interactions, and utilized to assist online mask fusion. Our method achieves competitive results on ScanNet200 and SceneNN, highlighting the value of integrating the geometric priors from RFM.
Our method inherits the limitations from the utilized RFM. The performance degrades on very long sequences, as the RFM tends to accumulate errors in geometry.

\noindent\textbf{Acknowledgments.} HB was supported by the EPSRC Visual AI grant EP/T028572/1.
{
    \small
    \bibliographystyle{ieeenat_fullname}
    \bibliography{main}
}
\clearpage
\setcounter{page}{1}
\appendix
\setcounter{table}{0}
\renewcommand{\thetable}{A\arabic{table}}
\setcounter{figure}{0}
\renewcommand{\thefigure}{A\arabic{figure}}
\maketitlesupplementary

The supplementary materials provide 1) more implementation details (Sec.~\ref{sec:supp_implementation}); 2) A component analysis (Sec.~\ref{sec:speed}); 
3) More qualitative examples for visualization and comparison (Sec.~\ref{sec:supp_vis}). We will release the code and pretrained models.

\section{Additional Implementation Details}
\label{sec:supp_implementation}

\noindent\textbf{Query Index Map}. The Query Index Memory (QIM) introduced in Sec.~3.4 relies on query index map $\mathcal{M}$, which is a mapping between a set of spatial keys and their associated query indices.

\noindent\textit{Map Update.} To ensure computational efficiency, we dynamically update the map to maintain a sparse key structure. Let $\mathcal{M}_{t-1}\in \{0,1\}^{n^k_{prev}\times n^q_{prev}}$ denote the map storing $n^q_{prev}$ previous contextual queries at $n^k_{prev}$ existing spatial keys. After receiving new queries and updating the global query bank $\mathcal{Q}_t\in \mathbb{R}^{(n^q_{prev}+n^q_{t})\times d}$, we first extend the dimension of $\mathcal{M}_{t-1}$ to $\{0,1\}^{n^k_{prev}\times(n_{prev}^q+n_t^q)}$ by padding with zeros. Then, we compute the Euclidean distance between the incoming spatial keys and the existing keys. Inspired by~\cite{wu2025point3r}, if an existing spatial key lies within a distance threshold $\delta_{\mathcal{M}}$ of one or more new keys, we update its position to the centroid of the matched new keys, and take the union of all corresponding query indices. Otherwise, new spatial keys that do not match any existing key are appended as new rows in $\mathcal{M}_t$.

\noindent\textit{Map Rasterization.} We render the query index map as a point cloud using the PyTorch3D point rasterizer~\cite{ravi2020pytorch3d}. To model the occlusions, we strictly rasterize the nearest point at each pixel.

\noindent\textbf{Other Implementation Details.} For the query index map described above and in Sec.~3.4, we set the distance threshold $\delta_{\mathcal{M}}$ to 0.3 for merging spatial keys. As shown in Sec.~4.1, \method is trained on ScanNet image data. After training, we directly test the model on ScanNet200 and SceneNN.  

\begin{table}[!t]
    \centering


    {
    \begin{tabular}{l|c}
    \specialrule{1.0pt}{0pt}{0pt}
    {\textbf{Component}} & Time (ms)\\
    \midrule
    Feature Extraction (DINO) & 6 \\
    Query Refinement &  18\\
    Memory Retrieval & 2\\
    Memory Update & 4\\
    Intra-frame Merging & 3\\
    Cross-frame Association & 22\\
    \midrule
    Total & 55 \\
         
\specialrule{1.0pt}{0pt}{0pt}
    \end{tabular}
    }
    \caption{\textbf{Component Analysis.} The time cost of individual component in \method.  
    }
\label{tab:speed}
\end{table}

\begin{figure*}[!t]
    \centering
    \includegraphics[width=\textwidth]{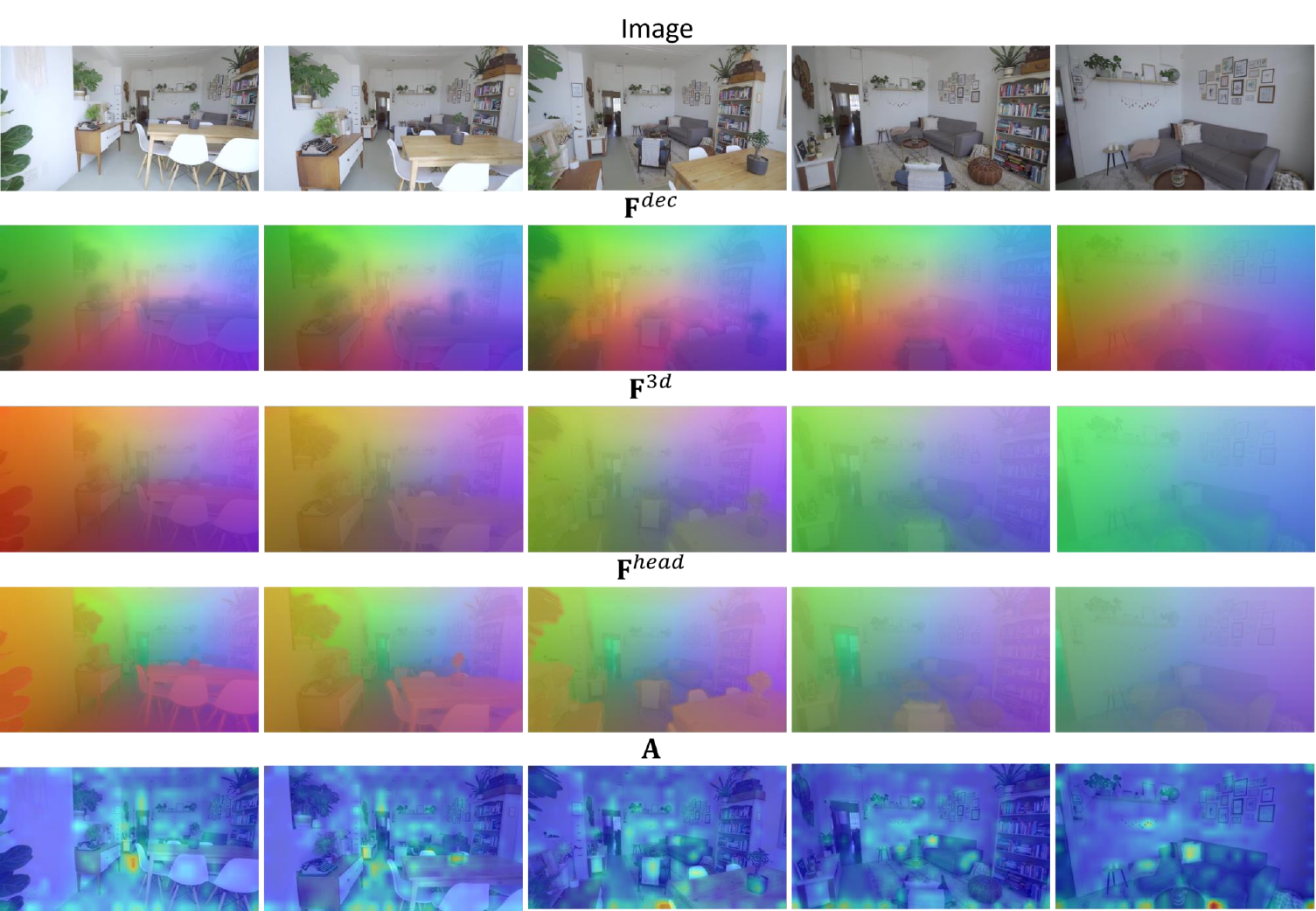}
    \caption{\textbf{Intermediate Representation Visualization}. We visualize the intermediate representations within CUT3R, specifically decoder features $\mathbf{F}^{dec}$, modulation features $\mathbf{F}^{mod}$, head features $\mathbf{F}^{head}$ and state attention maps $\mathbf{A}$ (from top to bottom). The decoder features $\mathbf{F}^{dec}$ contain local depth information. $\mathbf{F}^{mod}$ demonstrates world-coordinate consistency, as the modulation layer performs an implicit rigid transformation conditioned on the pose token. The head features $\mathbf{F}^{head}$ exhibit the sharpest geometric details and boundaries, since they are directly supervised by the pointmap regression loss. The state attention $\mathbf{A}$ consistently highlights specific semantic instances (e.g., pillow, plant), but remains spatially sparse.}
    \label{fig:state_pre}
\end{figure*}

\section{Component Analysis}
\label{sec:speed}
\subsection{Speed Analysis}

\textbf{\method}. The total inference time for \method is 321 ms per frame, including 121 ms for reconstruction and segmentation and 200 ms for 2D VFM mask generation, which corresponds to approximately 3.1 FPS. Excluding the RFM CUT3R and VFM CropFormer, our segmentation pipeline requires only 55 ms. Since the feature refinement network $\psi$ is used to obtain reference features for training in Eq.~5 and Eq.~6, it is not involved in the inference process. The breakdown of the 55 ms pipeline is as follows:

\noindent\textit{Feature Extraction}. We extract 2D semantic features $\mathbf{F}^{2d}$ using DINO in a single feed-forward step for 6 ms.

\noindent\textit{Query Refinement}. This module enhances query features by fusing information from the current frame and the retrieved memory context. This process consumes 18 ms due to attention mechanisms and feature fusion operations.

\noindent\textit{Memory Retrieval}. The retrieval step of the memory involves rasterizing the query index map $\mathcal{M}$ and indexing the global query bank $\mathcal{Q}$. Thanks to the highly efficient point rasterization algorithm and the sparsity of spatial keys, the rasterization process takes only 2 ms, which brings negligible time cost and allows for efficient contextual injection.

\noindent\textit{Memory Update}. The memory update process is relatively more computational intensive compared to retrieval, as it performs extensive search over all existing keys to merge/extend each new spatial key, in order to maintain map sparsity.

\noindent\textit{Intra-frame Merging}. As detailed in Sec.~3.5, to resolve the problem of oversegmentation from 2D VFM, we merge masks within the current frame based on the query feature map.

\noindent\textit{Cross-frame Association}. After intra-frame merging, we associate the merged masks from the current frame with the existing masks via bipartite matching. This uses a combined metric of query similarity, bounding box IoU and our novel state distribution token similarity. As the video frame increases, the number of tracked instances grows, resulting in an average processing time of 22 ms.

\noindent\textbf{OnlineAnySeg}. Previous work OnlineAnySeg consists of feature extraction, VoxelHashing updating and MaskGraph clustering. The workflow is more time-consuming than the reported results in their paper~\cite{tang2025onlineanyseg} because the time-consuming feature extraction and graph traversing procedure is not fully covered, which has been pointed out in a follow-up work~\cite{lan2025boxfusion} and a {Github issue}~\cite{onlineanyseg_issue}. For instance, to extract per-instance features, OnlineAnySeg generates multi-scale crops for each instance  and processes them individually with CLIP~\cite{radford2021icml}, which can take over 1 second for a single keyframe. 

With the 200 ms cost of the 2D VFM CropFormer~\cite{qi2023iccv} included, OnlineAnySeg requires a total of 3200 ms per frame. In contrast, our method, \method, takes only 321 ms, making it approximately 10 times faster even while performing simultaneous reconstruction and segmentation.

\subsection{Design Choice}

\textbf{Geometric Features}. In Sec.~3.1, we extract geometric features $\mathbf{F}^{3d}$ from CUT3R. To differentiate from the 2D semantic information $\mathbf{F}^{2d}$, $\mathbf{F}^{3d}$ must encode 3D spatial structure in world coordinates. We analyze different feature sources from CUT3R and visualize them in Fig.~\ref{fig:state_pre}. 

\noindent\textit{Decoder Features} $\mathbf{F}^{dec}$. Features from the image decoder $\mathcal{D}_I$ show similar color pattern in the image plane even as the camera moves, as it primarily contains depth information in the camera's self coordinate system.

\noindent\textit{Modulation Features} $\mathbf{F}^{mod}$. These features are extracted from the modulation layer, which is between the decoder and the world-coordinate prediction head. This layer uses self-attention conditioned on the pose token to perform an implicit rigid transformation of $\mathbf{F}^{dec}$ into world coordinates. Consequently, $\mathbf{F}^{mod}$ show consistent patterns that vary correctly with the changing 3D position.

\begin{figure}[!ht]
    \centering
    \includegraphics[width=\linewidth]{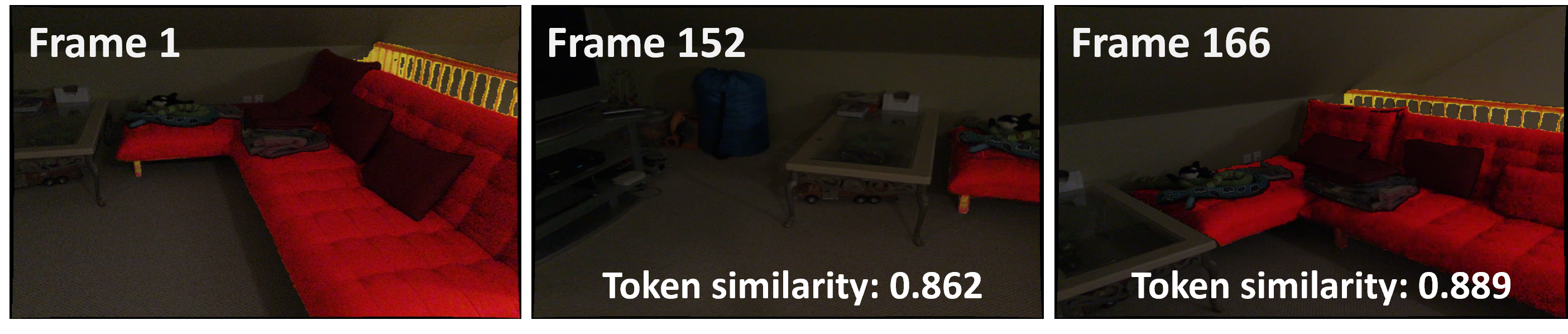}
\caption{ 
State distribution similarity in a long loop.
}
    \label{fig:supp_state}
\end{figure}

\begin{figure*}[!ht]
    \centering
    \includegraphics[width=\textwidth]{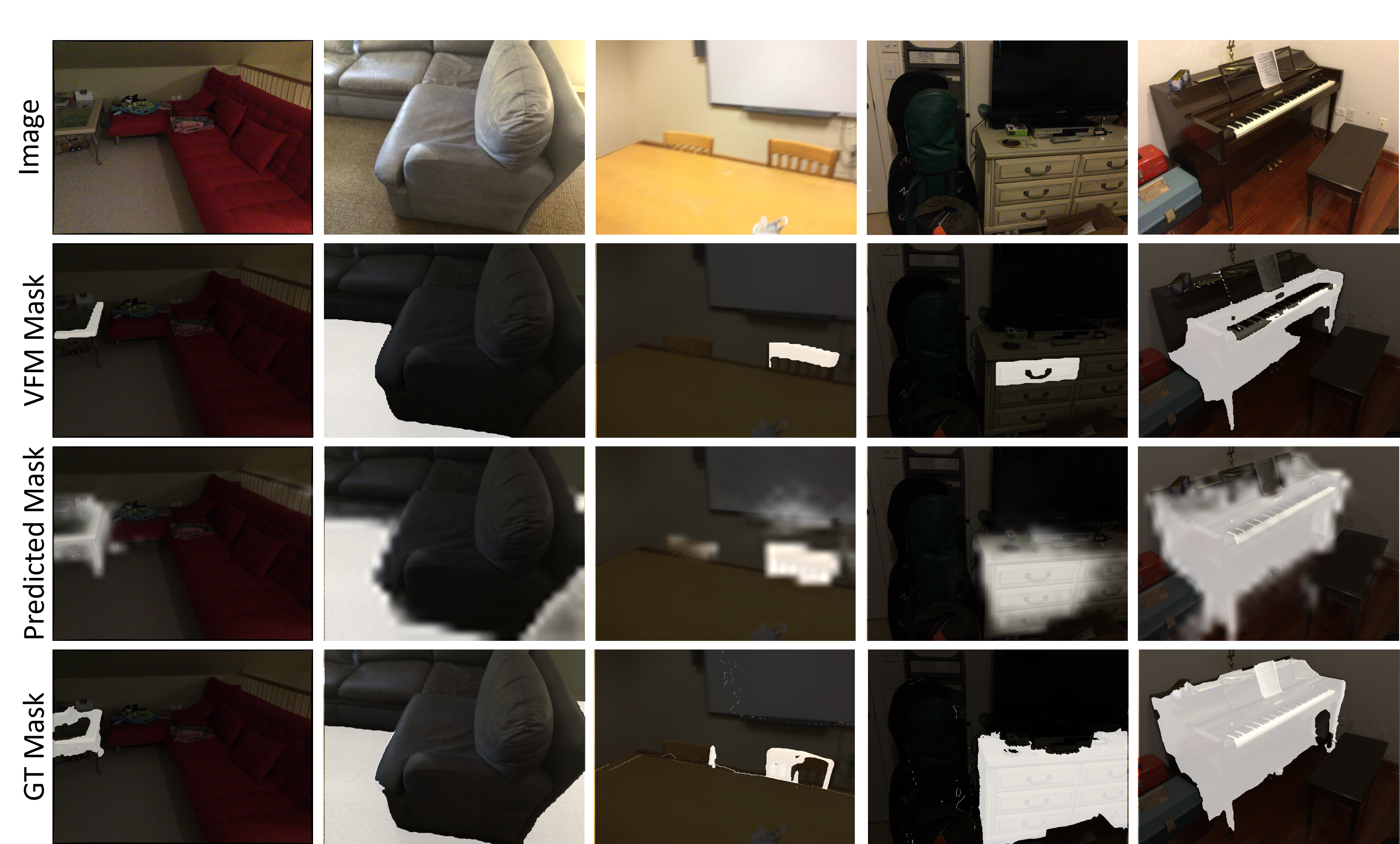}
    \caption{\textbf{Mask Prediction Visualization}. From top to bottom, we respectively visualize the original images, the input VFM masks, the predicted masks and the ground truth instance masks. While the VFM mask represents part of an oversegmented instance, its query correctly segments the whole instance in the predicted mask, validating that the self-supervised query learning and spatial-semantic distillation strategy mitigates the oversegmentation problem in 2D VFMs.}
    \label{fig:results_mask}
\end{figure*}

\begin{figure*}
    \centering
    \includegraphics[width=0.9\textwidth]{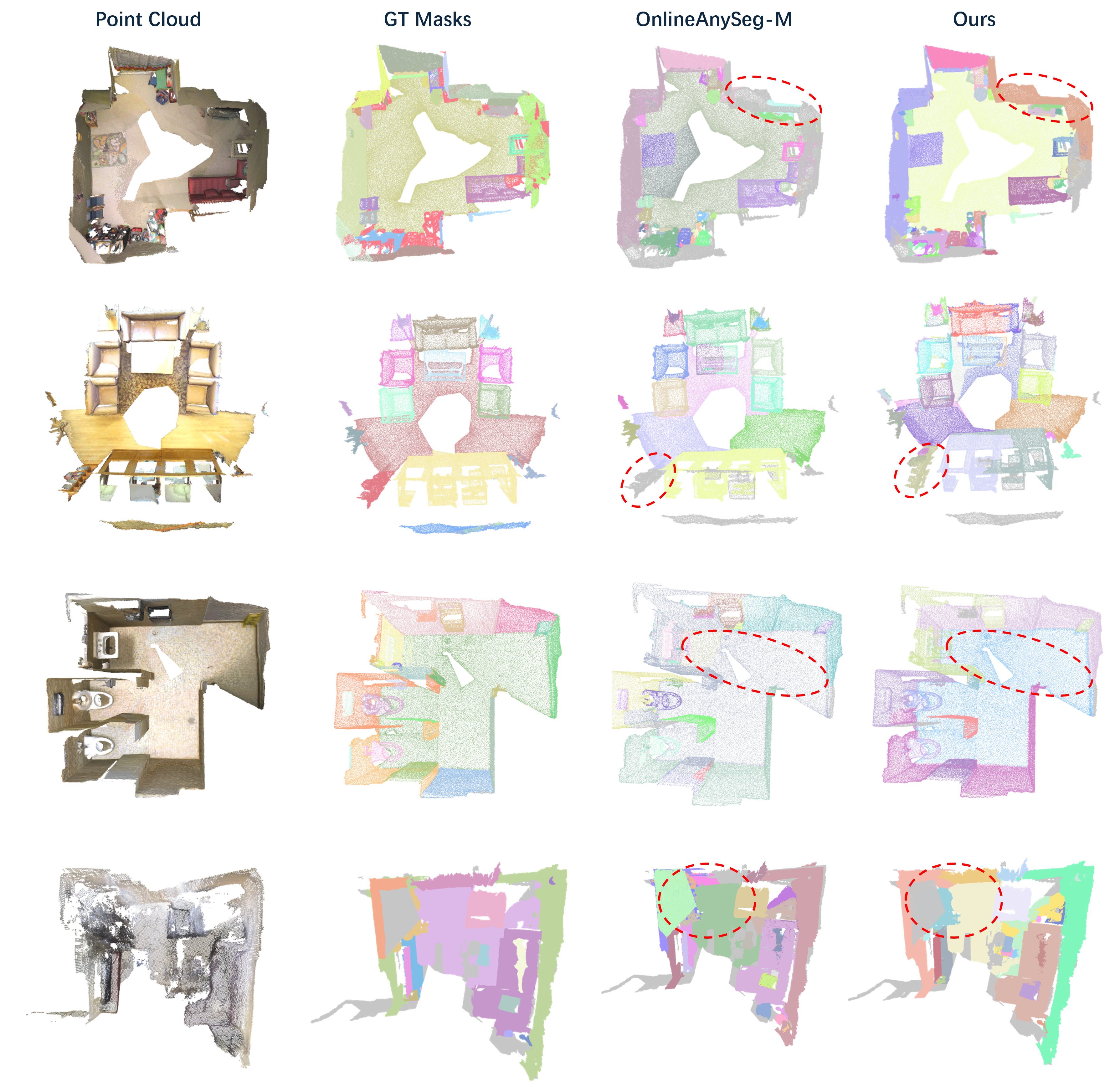}
    \caption{\textbf{Qualitative Comparison.} Qualitative examples of OnlineAnySeg-M and our method on ScanNet200 and SceneNN sequences. These results visually demonstrate that \method achieves superior instance segmentation. In contrast, OnlineAnySeg-M tends to fail in associating masks, which leaves significant unsegmented areas, as shown in the red dashed circles. The segmentation results are unprojected to ground truth point cloud for visualization. }
    \label{fig:supp_results}
\end{figure*}

\noindent\textit{Head Features} $\mathbf{F}^{head}$. CUT3R uses a DPT head to predict world pointmaps. The features $\mathbf{F}^{head}$ taken immediately before the final prediction layer contain the most explicit 3D information. Compared to $\mathbf{F}^{mod}$, $\mathbf{F}^{head}$ exhibits more distinct geometric boundaries and fine-grained spatial details, as it is directly supervised by the pointmap regression loss.

Despite $\mathbf{F}^{mod}$ and $\mathbf{F}^{head}$ both showing spatially varied geometric structure in world coordinates, we empirically find that $\mathbf{F}^{mod}$ provides better stability during training. Our hypothesis is that $\mathbf{F}^{head}$ is highly sensitive to high-frequency geometric noise due to the regression supervision, whereas $\mathbf{F}^{mod}$ offers a smoother and more robust representation. Therefore, we select $\mathbf{F}^{mod}$ as our geometric feature $\mathbf{F}^{3d}$.

\noindent\textbf{State Distribution Token}. In the last row of Fig.~\ref{fig:state_pre}, we visualize the attention map that state representation allocates to the image. The visualizations convey that, though the state consistently attends to specific objects (\emph{e.g.}, the plant pot on the table, the pillow on the sofa), the attention distribution is generally sparse, making it challenging to directly obtain pixel-wise association across frames. In Sec.~3.5, our proposed state distribution token aggregates this sparse signal into a robust instance-level representation. Our intuition is that the subset of state tokens responsible for encoding a specific instance region can be more stable and consistent. By performing a spatial masked summation over state attention map, we effectively identify the state tokens with high attention scores as the ones responsible for tracking the object within the mask, which produces a consistent state distribution token that successfully mitigates the correspondence ambiguity for inference-time association. 

\noindent\textit{State Distribution Consistency}. In Fig.~\ref{fig:supp_state}, we show the same sofa in a long loop of 166 keyframes. Although it leaves the view after frame 18 and reappears at frame 150,
its masks are still correctly matched via the highest state distribution similarity among all instances.

\section{Visualization}
\label{sec:supp_vis}

\noindent\textbf{Mask Prediction Visualization}. In Fig.~\ref{fig:results_mask}, we show examples of the predicted masks from \method with the input VFM masks. When VFM oversegments an object, our method consistently recovers the complete instance mask. This illustrates that the updated queries successfully capture comprehensive instance-level information, and the reference features have been effectively refined to be instance-discriminative, demonstrating the efficacy of our proposed self-supervised learning framework and spatial-semantic distillation strategy (see Sec.~3.3). Additionally, this property is critical for effective intra-frame merging in Sec.~3.5, as it ensures that queries corresponding to different parts of the same instance are mutually similar.



\noindent\textbf{Qualitative examples}.
More qualitative examples of our method on ScanNet200 and SceneNN are provided in Fig.~\ref{fig:supp_results}. Consistent with Fig.~3, our method merges masks more robustly on predicted geometry from CUT3R than OnlineAnySeg-M, which highly depends on precise geometry information to construct its VoxelHashing representation.


\end{document}